\documentclass{article} 
\PassOptionsToPackage{numbers,sort&compress}{natbib}
\usepackage{iclr2027_conference,times}
\AtBeginDocument{%
  \setcitestyle{numbers,square,comma,sort&compress}%
}

\usepackage[pagebackref]{hyperref}
\usepackage{url}
\usepackage[dvipsnames]{xcolor}
\usepackage{colortbl}
\usepackage{booktabs} 
\usepackage{multirow}
\usepackage{graphicx}
\usepackage{wrapfig}
\usepackage[most]{tcolorbox}
\usepackage{subcaption}
\usepackage[export]{adjustbox}
\usepackage{marvosym}

\definecolor{paperaccent}{HTML}{AC3239}

\newtcolorbox{paperbox}[1]{
    enhanced,
    breakable,
    width=\linewidth,
    colback=paperaccent!5!white,
    colframe=paperaccent!20!white,
    boxrule=0.4pt,
    borderline west={3pt}{0pt}{paperaccent},
    arc=2mm,
    outer arc=2mm,
    left=1mm,
    right=1mm,
    top=1mm,
    bottom=1mm,
    before skip=7pt,
    after skip=7pt,
    before upper={\textbf{#1}\par\smallskip},
}

\hypersetup{
    colorlinks = true,
    linkcolor = blue,
    anchorcolor = blue,
    citecolor = Maroon,
    filecolor = blue,
    urlcolor = blue,
    pdftitle = {Hypernetworks-for-Weight-Updates},
}

\usepackage{amsmath,amsfonts,bm}

\def\eqref#1{equation~\ref{#1}}

\def\1{\bm{1}}

\DeclareMathAlphabet{\mathsfit}{\encodingdefault}{\sfdefault}{m}{sl}
\SetMathAlphabet{\mathsfit}{bold}{\encodingdefault}{\sfdefault}{bx}{n}

\title{Learning to Predict Distributions over Weight Updates for Test-Time Adaptation}

\author{Azal Ahmad Khan\textsuperscript{1}\thanks{\Letter\ Correspondence: Azal Ahmad Khan \texttt{<khan1069@umn.edu>}}, Keshav Ramji\textsuperscript{2}, Tahira Naseem\textsuperscript{2}, Ali Anwar\textsuperscript{1}, Ramón Fernandez Astudillo\textsuperscript{2}\\
\textsuperscript{1}University of Minnesota\\ \textsuperscript{2}IBM Research AI\\
}

\begin{document}

\maketitle

\begin{abstract}
Hypernetworks have recently shown success in dynamically adapting the parameters of Large Language Models (LLMs) at runtime based on signals such as task descriptions or additional demonstrations. Here we ask: how much adaptation signal can be obtained using only the input query to an LLM? To answer this, we study query-conditioned hypernetworks for LoRA estimation. Further, we introduce distributional hypernetworks, able to produce not only point estimates of parameter adapters, but also a distribution over possible LoRAs. For this we propose a simple end-to-end loss using a differentiable Monte Carlo approximation and explore multiple distribution parameterizations including regression and convex combination variants. Results show that even using the mean of the learned distribution can outperform deterministic hypernetworks.
Crucially, the learned distribution enables a different form of test-time scaling: instead of spending additional compute only by sampling more token sequences from a fixed model, we sample weight updates, yielding multiple adapted models for the same query. Performance improves as more weight samples are considered and remains stronger than corresponding token-sampling adaptation baselines. Finally, we find that generated updates can transfer across queries, suggesting that the hypernetwork learns reusable structure in how the model should adapt. Together, these results show that query-conditioned distributions over weight updates can support both adaptation and test-time scaling.
\end{abstract}

\section{Introduction}
\label{sec:Introduction}

Updating parameters in a Large Language Model (LLM) requires gradient-based optimization over multiple rounds with sufficient data. This is a brittle process that can lead to catastrophic forgetting, even when using parameter efficient adaptation such as LoRA~\citep{hu2022lora}.
Gradient-descent is therefore particularly unsuitable for altering parameters of models at test-time, where there is little or no supervision data available. Hypernetworks offer a promising alternative way to do this. A hypernetwork is a neural network that generates the weights of another network \citep{ha2016hypernetworks}.
Hypernetworks are trained offline using ordinary backpropagation, but at inference time, they directly predict weight updates for a target network.
The learned parameters of the hypernetwork can therefore be viewed as slow weights that learn how to produce fast weights for each new input~\citep{jurgen1992learning, ba2016using}. Unlike learned optimizers \citep{andrychowicz2016learning}, the hypernetwork does not require gradients at inference time. It turns parameter adaptation from a repeated optimization process into a direct prediction.

\begin{figure}
    \centering
    \includegraphics[width=0.9\linewidth]{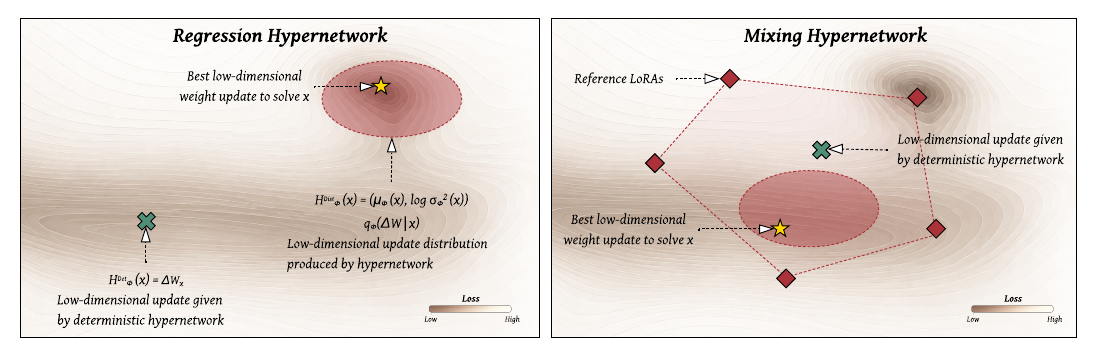}
    \vspace{-0.5em}
    \caption{\textbf{Distributional hypernetworks predict a query-conditioned distribution over LoRA updates.} The regression hypernetworks model the distribution directly in LoRA weight space, while the mixing hypernetwork models a distribution over combinations of reference LoRAs.}
    \vspace{-1em}
    \label{fig:main_figure}
\end{figure}

Recent work has shown success using hypernetworks to generate LoRA adapters from natural-language task descriptions \citep{charakorn2025text}, or generate adapters from multiple demonstrations of related examples~\citep{phang2023hypertuning}. These methods allow gradient-free adaptation, but still require such auxiliary information to be available. 
Here, we propose to predict a LoRA directly from the language model input. The key intuition is that different queries may reveal regularities in how the model should adapt, allowing the hypernetwork to learn an implicit mapping from input structure to useful parameter updates rather than relying on an explicit task description. This leads to a broader question: how much
adaptation signal is contained in a single query? Is it sufficient to determine a useful weight update? And if a single update is underdetermined, can the query instead constrain a useful distribution over plausible updates?
To address this, we propose a simple differentiable Monte Carlo approximation of the marginal log-likelihood using two types of updates: Gaussian LoRA regression and convex combinations of LoRA exemplars, depicted in Figure~\ref{fig:main_figure}. We explore multiple parameterization and optimization variants, such as weight correlations and loss aggregation, and compare our approach with a point-estimate hypernetwork.

Our experiments show a single query does indeed contain substantial adaptation
signal. It suffices to generate useful LoRA updates without task
descriptions or demonstrations, and can further support a distribution over
plausible updates.
The mean of the learned distribution outperforms deterministic hypernetworks by 4.6--11.7\% accuracy across 9 benchmarks. More importantly, this distribution becomes a new axis for test-time scaling: sampling different adapted models is competitive with, and often stronger than, sampling additional outputs from fixed weights. The generated updates also exhibit reusable structure across queries, as distributions produced from one subset of inputs can be aggregated and successfully applied to held-out queries without re-invoking the hypernetwork.

Our contributions can be summarized as follows:
\begin{itemize}
    \item We study how much adaptation signal can be extracted from a single query, showing that it is sufficient to generate useful query-conditioned LoRA updates without test-time optimization or task-level auxiliary information.
    \item We introduce \textbf{distributional hypernetworks} that predict a distribution over query-specific weight updates based on a simple differentiable Monte Carlo approximation.
    \item We introduce test-time scaling in weight space, where additional inference compute is used to sample alternative model updates rather than additional token sequences.
    \item We study the structure of the generated updates, including \textit{when} query-specific generation becomes useful and whether generated adaptations \textit{generalize} to other in-domain queries.
\end{itemize}

\section{Related Work}
\label{sec:related_works}


\paragraph{Hypernetworks.} Hypernetworks learn to generate the parameters of another network and provide a way to amortize adaptation without optimizing the target model at inference~\citep{ha2016hypernetworks}. In language models, HyperTuning~\citep{phang2023hypertuning} and HINT~\citep{ivison2023hint} generate parameter-efficient adaptations from demonstrations or instructions, while Text-to-LoRA (T2L;~\citep{charakorn2025text}) generates LoRA weights from a task description. More recent methods generate adaptations from longer contexts or related examples~\citep{liu2026shine, charakorn2026doctolora, volk2023example}. HyperDecoders move closer to instance-level adaptation by generating a separate adaptation from each input~\citep{ivison2022hyperdecoders}. These methods differ in what information is used to condition the hypernetwork, but predict a single adaptation. We show that a single query can provide enough information to generate useful query-conditioned LoRA updates and a distribution over possible LoRA updates. This contrasts with test-time training methods, which adapt the model through gradient-based optimization at inference time~\citep{sun2020test, wang2020tent, hu2025test, hu2025slot, tandon2025end, bansal2026let}.

\paragraph{Distributions over Weights.}
Learning distributions over neural network parameters has been studied in Bayesian neural networks and generative weight models. Bayesian Hypernetworks~\citep{krueger2017bayesian}, implicit weight uncertainty~\citep{pawlowski2017implicit}, and HyperGAN~\citep{ratzlaff2019hypergan} use learned generators to represent distributions over model parameters. These methods show that useful variation can be represented directly in weight space, but the resulting distributions are typically used for uncertainty estimation, ensembling, or representing a global model distribution.

More recent work studies distributions specifically in the low-rank adaptation space. ICM-LoRA~\citep{shao2025icmlora} and DiffLoRA~\citep{wu2024difflora} learn conditional generative models from collections of previously optimized adapters, while Bayesian LoRA methods model uncertainty around task-level adaptations~\citep{yang2023laplacelora,wang2024blob,zhang2025abmll}. LoRA ensembles similarly obtain multiple task-level adaptations through independent training~\citep{wang2023loraensembles}. Our distribution is different in both how it is produced and how it is used. It is conditioned on each query and trained directly through the frozen LM's response loss, rather than by reproducing previously optimized adapters or approximating a task-level posterior. Samples from the distribution are treated as alternative adaptations for the current query, allowing the hypernetwork to represent several useful updates rather than committing to a single one.

Further discussion of related work is deferred to Appendix~\ref{app:Full_related_works}.

\section{Method: Distributional Hypernetworks}
\label{sec:distributional_hypernets}

\subsection{Hypernetwork for LoRA Estimation}
\label{sec:det_hypernet}

We start with a pretrained autoregressive LM $p_{\theta}(y\mid x)$, parameterized by neural network weights $\theta$. Rather than modifying $\theta$ directly, a hypernetwork $\text{H}_\Phi$, with parameters $\Phi$, generates an input dependent weight update: 
\begin{equation}
\theta_x = \theta + \Delta\theta_x
\qquad\mbox{with}\qquad
\Delta\theta_x = \text{H}_\Phi(x),
\label{eq:hypernetwork_update}
\end{equation}
The hypernetwork is trained offline on a dataset $\mathcal{D}$ of query--response pairs, by maximizing the likelihood the adapted model assigns to the observed response $y$. We train the hypernetwork by
\begin{equation}
\Phi^{\star}
=
\arg\min_{\Phi} \hspace{2.5em}
\mathbb{E}_{(x,y)\sim\mathcal{D}}
\left[
-\log p_{\theta+\text{H}_\Phi(x)}(y\mid x)
\right]
\label{eq:hyp_out}
\end{equation}
During training, the loss is backpropagated through the LM and the generated update $\Delta\theta_x=\text{H}_\Phi(x)$, but only the hypernetwork parameters $\Phi$ are optimized, the pretrained parameters $\theta$ remain fixed. The resulting optimization problem is no different from conventional cross-entropy gradient descent over weights. The frozen LM provides the learning signal that teaches the hypernetwork which updates are useful for different queries. At test-time, only the query is available, and the trained hypernetwork directly produces $\Delta\theta_x$ in a single forward pass, after which the adapted model $p_{\theta+\Delta\theta_x}(y\mid x)$ generates the response. This procedure at test-time requires no ground-truth response, objective, backward pass, or iterative optimization. Consequently, the additional cost of adaptation is a single forward pass through $\text{H}_\Phi$, replacing per-query optimization with a learned mapping from queries to weight updates. 
The hypernetwork parameters $\Phi$ can be viewed as slow weights that learn to produce the query-specific fast weights $\Delta\theta_x$. 

Generating a dense update for every adapted weight matrix is impractical. For a matrix $\theta^{(\ell)} \in \mathbb{R}^{d_{\mathrm{out}} \times d_{\mathrm{in}}}$, a dense update contains $d_{\mathrm{out}}\times d_{\mathrm{in}}$ values. In a large transformer, this would require the hypernetwork to produce billions of values per query, making its output space prohibitively large. Previous work has shown that useful updates are low-dimensional~\citep{aghajanyan2021intrinsic, hu2022lora} such as LoRA. For a rank $r \ll \min(d_{\mathrm{in}},d_{\mathrm{out}})$ and scaling constant $\alpha$, the hypernetwork generates two query-specific factors, $\text{A}_x^{(\ell)} \in \mathbb{R}^{r \times d_{\mathrm{in}}}$ and $\text{B}_x^{(\ell)} \in \mathbb{R}^{d_{\mathrm{out}} \times r}$. Together, these factors induce a low-rank update. The number of values generated for each adapted matrix is therefore reduced from $d_{\mathrm{out}}d_{\mathrm{in}}$ to $r(d_{\mathrm{in}}+d_{\mathrm{out}})$. 
%
Let $\mathcal{M}$ denote the set of LM matrices adapted with LoRA. The hypernetwork outputs the LoRA factors for all matrices in $\mathcal{M}$, and these factors collectively define the effective update:
\begin{equation}
\Delta\theta_x^{(\ell)}
=
\frac{\alpha}{r}
\text{B}_x^{(\ell)}\text{A}_x^{(\ell)}.
\label{eq:update}
\end{equation}

Given a query $x$, a hypernetwork $\text{H}_\Phi(x)$ runs the input through $p_\theta$ to obtain the sequence of final-layer hidden states $\text{Z}_x=\text{F}_{\theta}^{\mathrm{hid}}(x)$ that are fed to the network. For each target matrix $s=(\ell,m)\in\mathcal{M}$, the cross-attention-based hypernetwork is conditioned on $\text{Z}_x$, a learned embedding for layer $\ell$, and a learned embedding for module type $m$. It produces the corresponding LoRA factors and thereby defines a deterministic query-specific update:
\begin{equation}
\Delta\theta_x^{(\ell)}=
\text{H}_\Phi\left(
\text{Z}_x,
e_{\mathrm{layer}}^{(\ell)},
e_{\mathrm{module}}^{(m)}
\right)
\label{eq:hidden}
\end{equation}
The hypernetwork is trained using the offline LM objective in Equation~\ref{eq:hyp_out}. Unlike prior methods that condition on a task description, demonstrations, or an explicit task identifier~\citep{charakorn2025text, phang2023hypertuning, ivison2022hyperdecoders}, our model derives its semantic conditioning entirely from the query. Furthermore, it uses the full sequence rather than a mean-pooled representation, since pooling can discard token-level information that may be useful for constructing the update. The layer and module embeddings identify where an update will be applied, but provide no information about the task itself. Consequently, the model requires no task-level metadata and may generate a different LoRA update for every query. This deterministic formulation produces a single point in the low-rank update space, which we generalize to a query-conditioned distribution in the following subsection.

\subsection{Distributional Hypernetwork}
\label{sec:distributional}
Predicting a single useful update from a query alone is an underdetermined problem. Several different adaptations may lead to a good response, and the query may not contain enough information to identify which one is best. A deterministic hypernetwork must nevertheless commit to one point in the LoRA parameter space, and when several solutions are plausible this point may be an average that resembles none of them. Recent evidence further suggests that the neighborhoods around large pretrained models are densely populated by diverse task-improving specialists~\citep{gan2026neural}. Here we propose to train a hypernetwork to predict a distribution over query-specific adapters.
\begin{equation}
\Phi^{\star}
=
\arg\min_{\Phi} \hspace{2.5em}
\mathbb{E}_{(x,y)\sim\mathcal{D}}
\left[\mathbb{E}_{\Delta\theta \sim q_\Phi(\omega\mid x)}
\left[
-\log p_{\theta+\Delta\theta}(y\mid x)
\right]\right]
\label{eq:dhyp_out}
\end{equation}
Since $\Delta\theta$ is continuous, we can optimize this loss directly using the reparameterization trick~\citep{kingma2013auto, rezende2014stochastic}:
\begin{equation}
\Phi^{\star}
=
\arg\min_{\Phi} \hspace{2.5em}
\mathbb{E}_{(x,y)\sim\mathcal{D}}
\left[\mathbb{E}_{\epsilon \sim \mathcal{N}_{n}(0, I_n)}
\left[
-\log p_{\theta+f_\Phi(x, \epsilon)}(y\mid x)
\right]\right]
\label{eq:hyp_out2}
\end{equation}
where $f_\Phi(x, \epsilon)$ transforms an initial parameterless $n$-dimensional Gaussian into some other distribution parametrized by $\Phi$.

At inference time, one option is to decode using the expected LoRA update:
\begin{equation}
y^{\star}
=
\arg\max_{y} \hspace{2.5em}
\log p_{\theta+\mathbb{E}_{q_\Phi}\left[\Delta\theta\mid x\right]}(y\mid x).
\label{eq:mean-adapter-decoding}
\end{equation}
Alternatively, we can perform minimum Bayes risk decoding (MBR) by sampling weights rather than tokens. We first sample $K$ adapters and greedily decode from each adapted model:
\begin{equation}
\begin{aligned}
\epsilon^{(k)} &\sim \mathcal{N}_n(0,I_n),
\qquad
\Delta\theta_x^{(k)} = f_\Phi(x,\epsilon^{(k)}),\\
\widehat{y}_x^{(k)}
&=
\arg\max_y \log p_{\theta+\Delta\theta_x^{(k)}}(y\mid x),
\qquad k=1,\ldots,K.
\end{aligned}
\label{eq:weight-sampled-hypotheses}
\end{equation}
Given the candidate set $\mathcal{Y}_x=\{\widehat{y}_x^{(1)},\ldots,\widehat{y}_x^{(K)}\}$ and a task-specific risk $R$, the MBR prediction is
\begin{equation}
y_{\mathrm{MBR}}^{\star}
=
\arg\min_{\widehat{y}\in\mathcal{Y}_x}
\frac{1}{K}\sum_{k=1}^{K}
R\!\left(\widehat{y},\widehat{y}_x^{(k)}\right).
\label{eq:weight-sampled-mbr}
\end{equation}
For example, majority voting is recovered by choosing the zero-one disagreement risk $R(\widehat{y},\widehat{y}')=\mathbb{I}[\widehat{y}\neq\widehat{y}']$. Each adapted model can therefore be decoded greedily, since predictive diversity comes from sampling model weights rather than tokens.


\paragraph{Gaussian LoRA Distribution}
Our first parameterization assumes Gaussian distributed LoRA factors. For a general correlated $n$-dimensional Gaussian, the reparameterized sample is
\begin{equation}
  f_\Phi(x,\epsilon)
  = \mu_\Phi(x) + L_\Phi(x)\epsilon,
  \qquad
  \epsilon \sim \mathcal{N}_n(0,I_n),
  \qquad
  \Sigma_\Phi(x)=L_\Phi(x)L_\Phi(x)^\top.
  \label{eq:correlated-gaussian}
\end{equation}

For each target matrix $s\in\mathcal{M}$, the hypernetwork of Equation~\ref{eq:hidden} uses one output head to predict the mean $\mu_\Phi^{(s)}(x)$ and another to predict the entries of the lower-triangular Cholesky factor $L_\Phi^{(s)}(x)$. The strictly lower-triangular entries are unconstrained, while positivity of the diagonal is enforced by applying a softplus transformation to the corresponding unconstrained outputs. This construction ensures that $f_\Phi(x,\epsilon)$ has mean $\mu_\Phi(x)$ and covariance $\Sigma_\Phi(x)$. A full covariance requires the hypernetwork to generate $n(n+1)/2$ Cholesky parameters, which is prohibitively expensive for the high-dimensional LoRA factor space. We therefore explore low-rank factorizations of the covariance as well as a diagonal covariance.
\begin{equation}
f_\Phi(x,\epsilon)
=
\mu_\Phi(x)+\sigma_\Phi(x)\odot\epsilon,
\qquad
\epsilon\sim\mathcal{N}_n(0,I_n).
\label{eq:diagonal-reparameterization}
\end{equation}

The above yields the following distribution over LoRA weights:
\begin{equation}
q_\Phi(\omega\mid x)
=
\mathcal{N}\bigl(
\omega;\mu_\Phi(x),\Sigma_\Phi(x)
\bigr),
\qquad
\Sigma_\Phi(x)=L_\Phi(x)L_\Phi(x)^\top.
\label{eq:sampling}
\end{equation}
For the diagonal specialization, the number of covariance outputs is linear rather than quadratic in $d_\Delta$. In this case the noise on $\text{A}$ and $\text{B}$ is independent and zero-mean, so the induced update satisfies $\mathbb{E}\bigl[\Delta\theta_x^{(s)}\bigr]=\frac{\alpha}{r}\mu_{\text{B},x}^{(s)}\mu_{\text{A},x}^{(s)}$. Thus, the product of the mean factors recovers a single deterministic adapter, while sampling explores the region around it.

\paragraph{Convex Hull LoRA Distribution}
A Gaussian over millions of LoRA parameters can place samples far from regions of weight space containing useful adapters. Our second parameterization instead constrains the search to combinations of a bank of $\text{R}$ trained reference adapters $\{\widehat{\Delta\theta}_j\}_{j=1}^{\text{R}}$. The hypernetwork predicts the mean $\mu_{z,\Phi}(x)$ and Cholesky factor $L_{z,\Phi}(x)$ of a Gaussian over mixing logits. Applying the softmax maps these logits to mixing weights on the simplex:
\begin{equation}
\begin{gathered}
\pi_\Phi(x,\epsilon)
=
\operatorname{softmax}\!\left(
\mu_{z,\Phi}(x)+L_{z,\Phi}(x)\epsilon
\right),
\qquad
\epsilon\sim\mathcal{N}_{\text{R}}(0,I_{\text{R}}),\\
f_\Phi(x,\epsilon)
=
\sum_{j=1}^{\text{R}}
\pi_{\Phi,j}(x,\epsilon)\widehat{\Delta\theta}_j.
\end{gathered}
\label{eq:mixture}
\end{equation}
We denote the resulting distribution over LoRA weights by
\begin{equation}
q_\Phi(\omega\mid x)
=
\operatorname{CH}\!\left(
\omega;
\mu_{z,\Phi}(x),
\Sigma_{z,\Phi}(x)
\right),
\qquad
\Sigma_{z,\Phi}(x)=L_{z,\Phi}(x)L_{z,\Phi}(x)^\top.
\label{eq:convex-hull-distribution}
\end{equation}
Here $\operatorname{CH}$ is notation we introduce rather than a standard distribution family. The mixing weights $\pi_\Phi(x,\epsilon)$ follow a logistic-normal distribution, and $\operatorname{CH}$ is its linear pushforward through the map $\pi\mapsto\sum_{j=1}^{\text{R}}\pi_j\widehat{\Delta\theta}_j$. Its support is therefore contained in the convex hull of the reference adapters, restricting the model to a low-dimensional region of weight space. The two parameterizations are not directly comparable in scale because the first operates in the $d_\Delta$-dimensional factor space, while the second operates in an $\text{R}$-dimensional logit space with $\text{R}-1$ effective degrees of freedom.

\section{Distributional Hypernetwork Training}

For training the Distributional Hypernetwork, we use the Monte Carlo gradient estimate of Equation~\ref{eq:hyp_out2}. For a single example $(x_i,y_i)$ this is
\begin{equation}
\nabla_\Phi\mathcal{L}_i(\Phi)
\approx
-\frac{1}{K}
\sum_{k=1}^{K}
\nabla_\Phi\log p_{\theta+f_\Phi(x_i,\epsilon_{ik})}(y_i\mid x_i).
\label{eq:per-example-gradient}
\end{equation}

Here $\epsilon_{i1},\ldots,\epsilon_{iK}\overset{\mathrm{iid}}{\sim}\mathcal{N}_{n}(0,I_n)$. In practice, we train with mini-batches $\mathcal{B}$ of query--response pairs and average these per-example gradients across the mini-batch.

In addition to approximating the reparameterized expectation through an average of $K$ samples, we also explore the $\max$ and $\min$ operators. Writing the sampled negative log-likelihood as $\ell_k=-\log p_{\theta+f_\Phi(x,\epsilon_k)}(y\mid x)$, these reductions are $\frac{1}{K}\sum_k\ell_k$, $\min_k\ell_k$, and $\max_k\ell_k$, respectively. The mean is the principled default: it is an unbiased Monte Carlo estimator of the expected-risk objective in Equation~\ref{eq:hyp_out2}, gives every sample a gradient, and connects to established training methods based on stochastic weight distributions~\citep{blundell2015weight,kingma2015variational}.

The minimum and maximum instead optimize $K$-dependent order statistics and are therefore biased with respect to that expected-risk objective. The minimum is an optimistic best-of-$K$ or hard-assignment objective: it selects the highest-likelihood sampled adapter and resembles multiple-choice learning, where specializing different hypotheses can improve coverage of ambiguous outputs~\citep{guzmanrivera2012multiple,lee2016stochastic}. This can align well with test-time sampling and selection, but its gradient depends on a single winning sample and can favor lucky samples or whichever mode wins early. The maximum is a pessimistic worst-of-$K$ objective related to robust optimization and adversarial weight perturbation~\citep{madry2018towards,foret2021sharpness}. It can suppress catastrophic samples and favor locally robust solutions, but because the covariance is learned, the model can reduce the worst-case loss simply by collapsing its distribution.

All three reductions serve different goals: the mean targets average sampled performance, the minimum rewards search coverage, and the maximum rewards robustness. Behavior of minimum and maximum changes strongly with $K$. Without an explicit entropy reward, variance constraint, or prior, the mean and especially the maximum tend to shrink the learned covariance, while the minimum may preserve spread without guaranteeing diversity. Further details are in Appendix~\ref{app:training_details}.

\section{Experiments}
\label{sec:experiments}
\subsection{Experimental Setup}
\label{sec:setup}

\begin{figure}[t]
    \centering
    \begin{subfigure}[t]{0.35\linewidth}
        \centering
        \includegraphics[width=\linewidth]{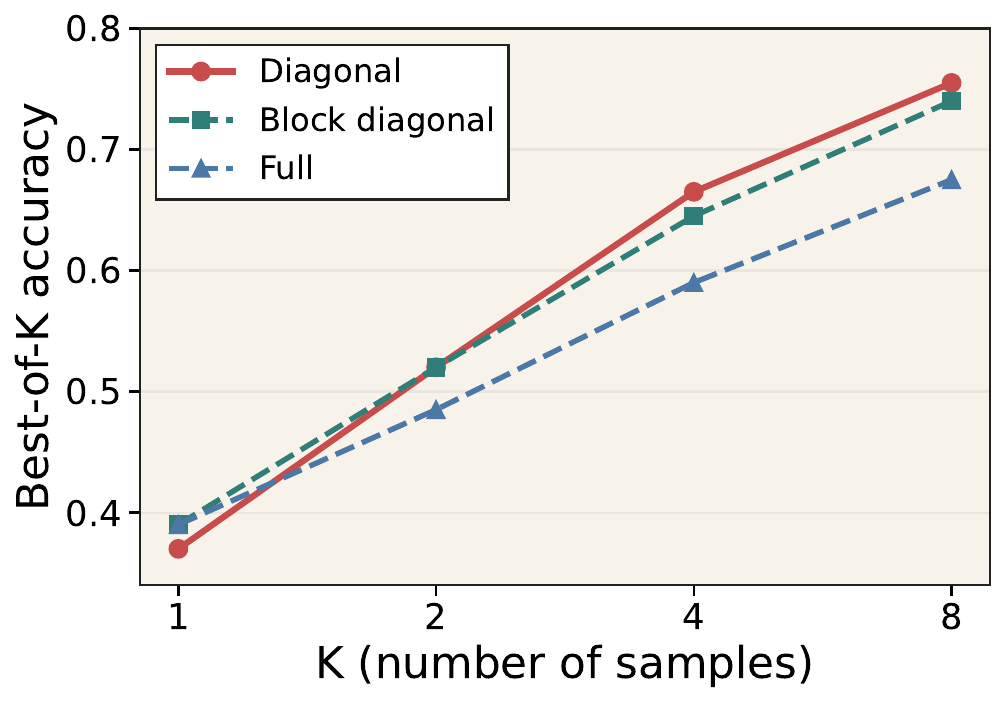}
        \label{fig:covariance_main}
    \end{subfigure}
    \begin{subfigure}[t]{0.35\linewidth}
        \centering
        \includegraphics[width=\linewidth]{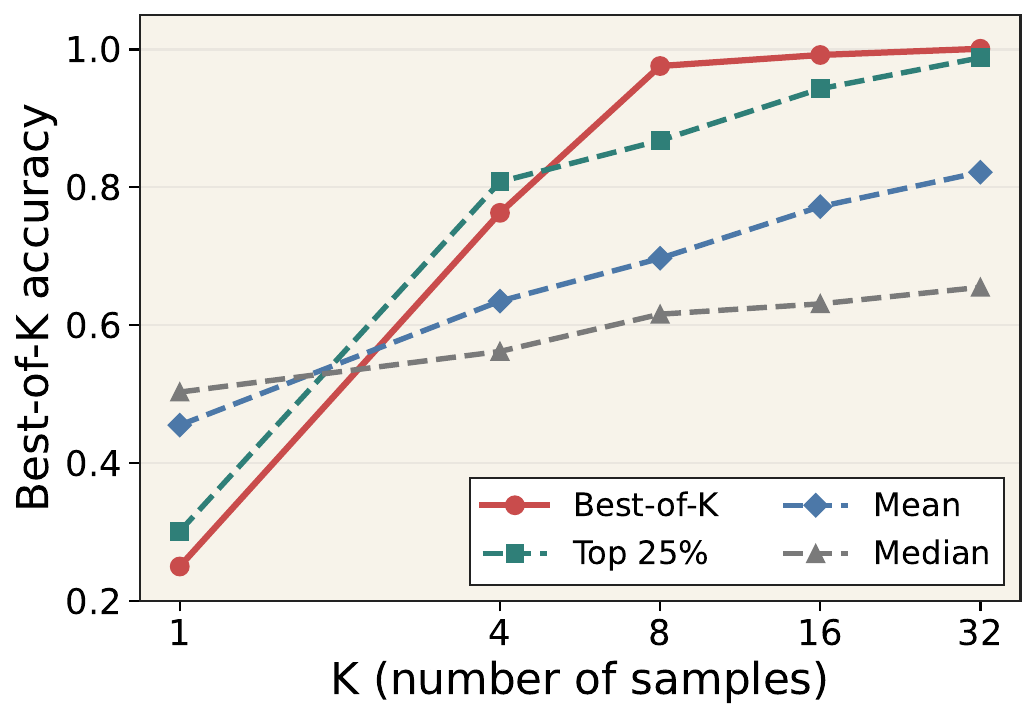}
        \label{fig:loss_main}
    \end{subfigure}
    \vspace{-1.5em}
    \caption{
    \textbf{Design choices for learning the weight distribution.}
    (a) Diagonal covariance performs comparably to more expressive covariance
    parameterizations.
    (b) Best-of-$K$ training preserves a more searchable distribution as the
    number of sampled updates increases.}
    \vspace{-1em}
    \label{fig:distribution_design}
\end{figure}

\paragraph{Models, training and evaluation data.}
We evaluate our method on five language models, Qwen3-1.7B, Qwen3-4B, Gemma3-1B-Instruct, Llama3.2-1B-Instruct, and Llama3.2-3B-Instruct. We use a diverse set of benchmarks, covering standard LLM tasks GSM8K, MATH500, HumanEval, ARC-Challenge, MMLU-STEM, MedQA and GPQA-Diamond. Hypernetwork and baselines are trained using another collection of tasks NuminaMath CoT, MATH train, MathInstruct, MedMCQA (using only its training split), Magicoder OSS Instruct 75K, and CAMEL AI. These datasets cover a broad set of problems from mathematics, coding, science, and medicine. The data results in approximately 1.3 million training examples containing around 530 million tokens. We will report macro accuracy, as the unweighted mean of accuracy across the seven evaluation benchmarks.

\paragraph{Distributional Hypernetwork architecture.}
We start by constructing a deterministic version of the hypernetwork from Section~\ref{sec:det_hypernet}, by modifying the T2L model~\citep{charakorn2025text}. This model receives as input a layer embedding, and a module embedding and a task embedding produced with a text encoder. We replace the last embedding by a learned pooled representation of the query, implemented with an additional attention mechanism over the last layer of the language model after it encodes the query. This network is trained with the loss in Equation~\ref{eq:dhyp_out}. We construct a sampler for the distributional version of this network as explained in Section~\ref{sec:distributional_hypernets}. This uses the same backbone but separately predicts mean and standard deviation. It then applies the Gaussian or convex Hull transformations to obtain samples from the LoRA distribution. 
The regression hypernetwork predicts LoRA factors directly, while the mixing hypernetwork predicts logits over a bank of reference LoRAs, which are softmaxed to obtain the mixture weights. The reference LoRAs are first trained for three epochs on the same training data, after which the hypernetwork is trained for one additional epoch. The mixing setting, therefore, receives more training exposure than the regression setting.

Training follows Equation~\ref{eq:hyp_out}. We explore training variants including Gaussian covariance learning as in Equation~\ref{eq:correlated-gaussian}, modeled with a low-rank matrix product for tractability, as well as average, best-of-K and worst-of-K losses. We observe empirically that diagonal covariance and best-of-K losses provide the best results (Figure~\ref{fig:distribution_design}). Additional details about these analyses can be found in Appendix~\ref{sec:learning_dist_app}. The resulting hypernetworks have approximately 300M parameters. We find that larger hypernetworks mainly improve single-update performance, while the gap narrows substantially as the weight-sampling budget increases (Appendix~\ref{app:hypernet_capacity}). Inference is done either using the sampler or the predicted mean LoRA, as detailed. We provide the full architecture and additional implementation details in Appendix~\ref{app:hypernet_details}.

\paragraph{Baselines.}
We compare the proposed method with Full FineTuning (FullFT) and standard LoRA. We also include Mixture-of-LoRA (MoL), which uses a learned router to select one LoRA from a fixed set using top-1 routing~\citep{feng2024mixture, zhao2024loraretriever}. MoL is also conditioned on the input for the routing operation. This allows us to separate the benefit of generating an update from the benefit of simply choosing among several learned updates. We also consider the deterministic version of the distributional hypernetwork above. All methods use the same rank 8 except for the rank sensitivity analysis.

\subsection{Predicting Weight Updates from Query}
\label{sec:distribution_improves_update}

\begin{figure}[t]
    \centering
    \begin{minipage}[t]{0.38\linewidth}
        \centering
        \includegraphics[
            width=\linewidth,
            valign=t
        ]{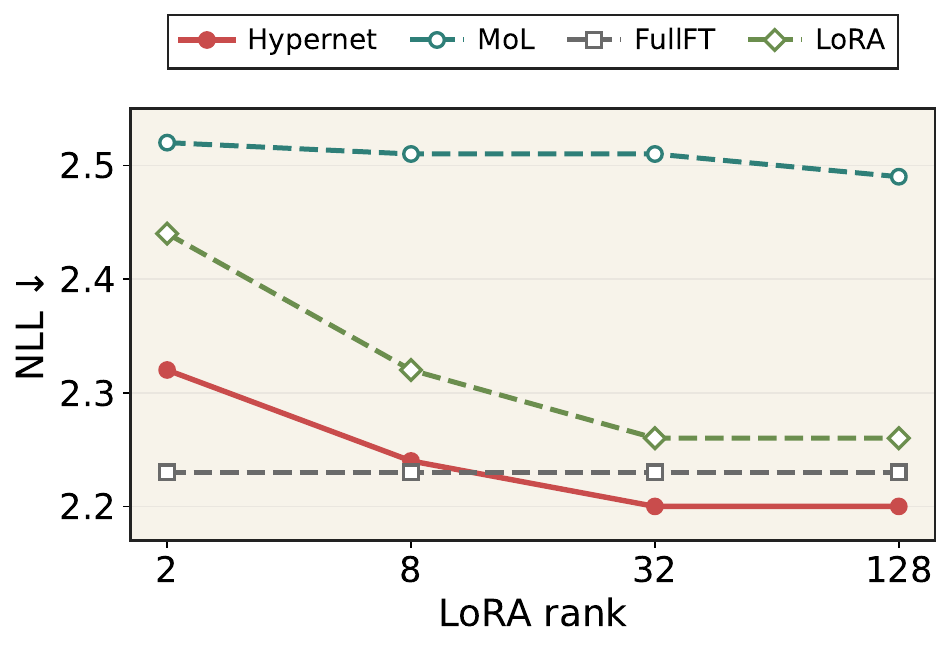}
    \end{minipage}
    \hfill
    \begin{minipage}[t]{0.58\linewidth}
        \vspace{0pt}
        \centering
        \footnotesize
        \setlength{\tabcolsep}{1.5pt}
        \renewcommand{\arraystretch}{1.05}

        \adjustbox{valign=t,max width=\linewidth}{%
        \begin{tabular}{lccccc}
        \toprule
        \textbf{Method}
        & \shortstack{\textbf{Gemma3}\\\textbf{1B}}
        & \shortstack{\textbf{Llama3.2}\\\textbf{1B}}
        & \shortstack{\textbf{Qwen3}\\\textbf{1.7B}}
        & \shortstack{\textbf{Llama3.2}\\\textbf{3B}}
        & \shortstack{\textbf{Qwen3}\\\textbf{4B}} \\
        \midrule
        FullFT                & 22.4 & 34.0 & 45.4 & 50.4 & 64.8 \\
        LoRA                  & 18.7 & 25.2 & 46.0 & 46.7 & 60.6 \\
        MoL                   & 21.0 & 22.6 & 42.2 & 44.1 & 60.6 \\
        \midrule
        Hypernet (reg.)       & 17.7 & 23.3 & 41.3 & 47.4 & 61.4 \\
        Dist. Hypernet mean (reg.) & 21.1 & 27.1 & 48.4 & 50.5 & 61.5 \\
        \midrule
        Hypernet (mix.)       & 19.5 & 36.3 & 50.1 & \textbf{56.6} & \textbf{67.1} \\
        Dist. Hypernet mean (mix.) & \textbf{31.3} & \textbf{38.9} & \textbf{55.7}
                              & 56.0 & 65.3 \\
        \bottomrule
        \end{tabular}
        }
    \end{minipage}
    \caption{
    \textbf{(a)} Effect of LoRA rank on adaptation performance. Increasing the rank improves the Hypernetwork performance over FullFT. \textbf{(b)} Macro accuracy over seven benchmarks across five base models. Distributional training generally improves the predicted update.}
    \vspace{-1.5em}
    \label{fig:hypernet_update_properties}
\end{figure}




We first compare FullFT, LoRA, and MoL against the deterministic and distributional hypernetworks using the same training data, with the mean estimate used for the distributional hypernetwork.

We find that the LoRAs generated by the hypernetwork show trends similar to standard LoRAs but generally perform better. Figure~\ref{fig:hypernet_update_properties} (left) shows average negative log-likelihood for the Qwen3-1.7B and Qwen3-4B models on held-out test splits of the FLAN tasks. Both standard and hypernetwork-generated LoRAs (regression) improve with rank. Hypernetwork-generated LoRAs even outperform full fine-tuning at rank $32$ or higher, while the generated LoRA updates remain smaller than full-model updates. Figure~\ref{fig:hypernet_update_properties} (right) shows average performance over the evaluation benchmarks for different models. As a general trend, hypernetworks outperform all other methods. The mixing hypernetwork performs better, which may reflect its larger size and the additional training of a bank of LoRAs. Parameter counts, including those of the hypernetwork itself, are reported in Appendix~\ref{app:hypernet_size}.

\begin{paperbox}{Finding 1: \textit{A Single Query Contains Useful Adaptation Signal}}
A single query contains enough adaptation signal to generate effective LoRA updates without task descriptions or demonstrations. The resulting updates match or outperform standard low-rank adaptation baselines.
\end{paperbox}

Another relevant trend is that the distributional hypernetwork outperforms its deterministic counterpart in all but the largest model sizes. We attribute this improvement in part to regularization introduced by sampling during training. A deterministic hypernetwork is optimized only at the update it predicts, while a distributional hypernetwork receives gradients through perturbed updates sampled around its prediction. The resulting objective therefore depends not only on the loss at one point, but also on nearby points in weight space. For zero mean Gaussian perturbations, the expected loss can be viewed locally as the loss at the mean together with a curvature dependent term, favoring solutions whose nearby perturbations also have low loss. In this sense, distributional training encourages the predicted update to lie in a more robust region of the adaptation space rather than optimizing a single point. An extended set of examples can be found in Appendix~\ref{app:dist_vs_det_full}.

\begin{paperbox}{Finding 2: \textit{Distributional Training Improves the Predicted Update}}
Training the hypernetwork as a distribution produces a stronger single update than directly training a point estimate. Even without weight sampling at test time, the mean of the learned distribution consistently outperforms the deterministic hypernet.
\end{paperbox}



\subsection{Predicting Weight Distribution from Query}
\label{sec:distributional_hypernet}

\begin{figure}[t]
    \centering
    \includegraphics[width=0.95\linewidth]{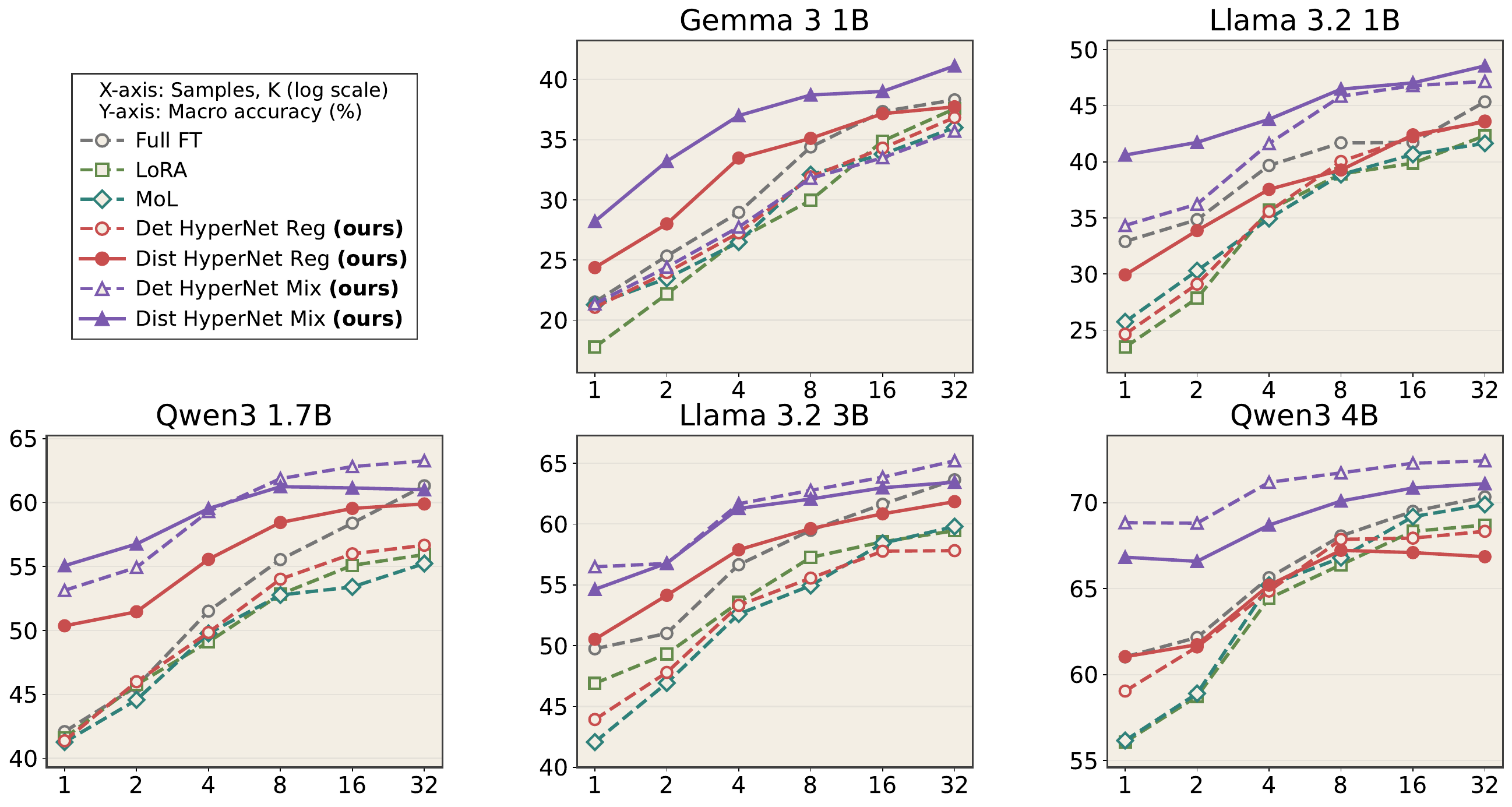}
    \caption{\textbf{Scaling test-time compute by sampling weights.} We report macro accuracy over seven benchmarks as the sampling budget increases. For each query, we generate $K$ candidate answers and use majority voting to obtain the final prediction. Dashed lines correspond to token sampling from a fixed model, including baselines and deterministic hypernetworks. Solid lines correspond to distributional hypernetworks, where each candidate is generated using a sampled weight update.}
    \vspace{-1em}
    \label{fig:weight_sampling}
\end{figure}



The previous section showed that a single query contains enough adaptation signal to predict a useful update. Here we ask whether the query can also enable learning a distribution over good LoRA updates. 
A distributional hypernetwork provides more than a stronger single update.
It also allows us to sample a set of alternative adaptations at
test time, thus giving an additional way to spend test-time compute. Most test-time scaling methods keep the model weights fixed and spend additional compute on sampling more output sequences~\citep{shao2024deepseekmath, snell2024scaling, muennighoff2025s1}. With the distributional hypernet, we can instead sample multiple LoRA updates, producing a different adapted model for each generation.

We compare these two ways of using additional compute through majority voting. For FullFT, LoRA, MoL, and deterministic hypernetworks, all $K$ generations are sampled from the same adapted weights. These methods therefore scale only through token sampling. For distributional hypernetworks, each generation uses a separately sampled weight update from the distribution. Increasing $K$ explores different adapted models in addition to output sequences. We use temperature $0.8$ for token sampling, while generations from sampled weight updates are decoded greedily ($T=0$).


Figure~\ref{fig:weight_sampling} shows the resulting scaling behavior across five language models using macro accuracy over seven benchmarks. Weight sampling remains competitive with token sampling as the test-time budget increases, and is often stronger than the corresponding deterministic hypernet. This is especially clear for Qwen3 1.7B and Gemma 3 1B, where sampling from the learned weight distribution provides substantial gains over repeatedly decoding from a single predicted update. Even when the advantage is smaller, as for Qwen3 4B, weight sampling remains competitive with strong adaptation baselines. These results suggest that useful test-time diversity can come not only from sampling different outputs, but also from sampling different versions of the adapted model itself. We report the full results in Appendix~\ref{app:full_test_time_scaling}.
We consider regression and mixing hypernetworks separately and do not compare their performances against each other. Our comparison instead focuses on the effect of deterministic token sampling and distributional weight sampling within each parameterization.

\begin{paperbox}{Finding 3: \textit{Test-Time Compute Can Be Spent in Weight Space}}
Sampling weight updates provides a new way for test-time scaling. Across model families, weight sampling remains competitive with token sampling and often performs better, showing that useful diversity can come from changing the adapter rather than only its generated tokens.
\end{paperbox}

\subsection{Reusing Generated Updates Across Queries}
\label{sec:cross_query_transfer}
\begin{figure}
    \centering
    \includegraphics[width=\linewidth]{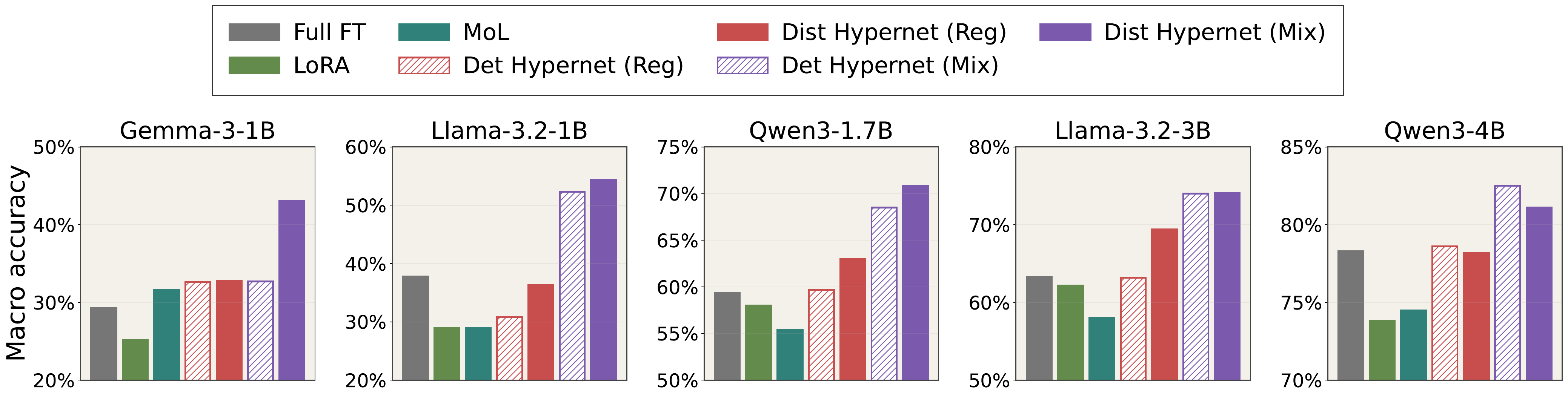}
    \vspace{-1em}
    \caption{Generated update distributions transfer across queries. We aggregate hypernetwork outputs from one half of each benchmark and apply the resulting adaptation to the held-out half.}
    \vspace{-0.5em}
    \label{fig:cross_query_transfer}
\end{figure}
So far, we have generated a separate update for every test query. We next ask whether the updates produced for different queries contain structure that can be accumulated and reused. For each benchmark, we split the examples into two disjoint halves. We use only the prompts from the first half to generate hypernetwork outputs and combine them into a single shared LoRA. This adapter is then kept fixed and used for every query in the second half. Importantly, the hypernetwork is never called on the held-out queries and no labels from the first half are used. 
Although the hypernetwork observes the first-half queries as conditioning inputs, neither it nor the baselines receives labels or undergoes optimization on these examples.
The experiment therefore measures whether the updates generated from one set of queries can transfer to unseen queries from the same benchmark.

For the regression hypernet, we obtain LoRA factors for each query in the first half. The deterministic hypernetwork contributes its prediction, while for the distributional hypernetwork we use the predicted mean without sampling. We average the $A$ factors across queries and separately average the $B$ factors, producing one rank-$r$ LoRA reused on the held-out half. For the mixing hypernet, each query produces mixing weights over the reference LoRA bank. We average these coefficients across the first half and use them to construct one shared adapter for the second half.

Figure~\ref{fig:cross_query_transfer} shows that the generated updates retain useful information across queries. Although the shared adapter is constructed entirely from a disjoint set of prompts, it remains effective on the held-out half, and the distributionally trained hypernetworks generally produce stronger reusable adaptations than their deterministic counterparts. This suggests that the hypernetwork is learning more than isolated per-query corrections: its outputs capture regularities that can be accumulated across examples and reused without calling the hypernetwork on every new query. Thus, the same mechanism can support both query-specific adaptation and a persistent shared adapter built from previously observed inputs.

\section{Conclusion}
We study how much adaptation signal can be extracted from a single query. Our results show that this is sufficient to generate useful query-conditioned LoRA updates and distributions over plausible updates. Distributional hypernetworks further improve the predicted update and enable test-time scaling through weight sampling. Sampling different adapted models is competitive with, and often stronger than, sampling more outputs from fixed weights. Finally, the generated updates exhibit reusable structure across queries, supporting both per-query and shared task-level adaptation.

\section{Limitations}
Our experiments focus on 1--4B parameter models and LoRA-based adaptation, and it remains to be seen how the behavior scales to substantially larger models. The hypernetwork must also be kept in memory and adds one forward pass per query, but inference remains considerably lighter than optimization-based adaptation since it requires no backward pass, optimizer state, or iterative updates. Given the consistent gains from generated and sampled updates, scaling this approach to larger models and more efficient hypernetworks is a promising direction for future work.

\section*{Reproducibility Statement}
We provide the information necessary to reproduce our experiments throughout the main paper and appendix. We specify the models and benchmarks used for evaluation, the datasets used for training, and the important training and implementation hyperparameters. Additional experimental and implementation details are provided in the appendix.

\section*{AI Use Statement}
We used AI tools to assist with writing and editing code and for editing portions of the manuscript. All AI-assisted code and text were reviewed by the authors. The authors take full responsibility for the final content of the paper, including all code, text, claims, and results.

\bibliographystyle{unsrtnat}
\bibliography{bibliography}

\clearpage
\onecolumn

\appendix

\section{Extended Related Work}
\label{app:Full_related_works}

\paragraph{Test-Time Training.}
Test-time training adapts a model to the current input by updating its parameters during inference. Early methods construct a self-supervised objective from the test input and take one or more gradient steps to adapt to distribution shift~\citep{sun2020test,wang2020tent}. More recent work applies the same idea to language models, including sample-specific optimization, test-time learning, and adaptation to long contexts~\citep{hu2025slot,hu2025test,tandon2025end,bansal2026let}. These methods differ in the objective and the parameters being updated, but all perform optimization after seeing the test input. Our goal is complementary. We train the adaptation mechanism beforehand so that a useful update can be produced directly from the query without gradient-based optimization at test time.

\paragraph{Gradient-Free Adaptation.}
Hypernetworks provide a natural way to amortize parameter adaptation by learning a network that generates the parameters of another network~\citep{ha2016hypernetworks}. This idea is closely related to fast weights, where a set of learned parameters controls another set of parameters that can change more rapidly~\citep{jurgen1992learning,ba2016using}. In language models, HyperTuning and HINT generate parameter-efficient adaptations from demonstrations or instructions~\citep{phang2023hypertuning,ivison2023hint}, while Text-to-LoRA generates a LoRA directly from a natural-language task description~\citep{charakorn2025text}. More recent work such as SHINE and Doc-to-LoRA maps longer contexts into LoRA weights in a single pass~\citep{liu2026shine,charakorn2026doctolora}, and example-based hypernetworks generate adaptations from examples from related domains~\citep{volk2023example}. HyperDecoders move closer to instance-level adaptation by generating a separate decoder adaptation from each input~\citep{ivison2022hyperdecoders}. These approaches show that useful parameter updates can be generated without test-time backpropagation. Our setting differs in that the semantic information comes only from the current query and the hypernetwork predicts a distribution over possible LoRA updates rather than a single adaptation.

\paragraph{Distributions over Weights.}
Learning distributions over neural network parameters has a long history in Bayesian neural networks and generative weight models. Bayes by Backpropagation learns distributions over weights through variational inference~\citep{blundell2015weight}, while Bayesian Hypernetworks, implicit weight uncertainty, and HyperGAN use learned generators to represent richer distributions over neural network parameters~\citep{krueger2017bayesian,pawlowski2017implicit,ratzlaff2019hypergan}. In these approaches, samples typically represent uncertainty or different models drawn from a global learned weight distribution.

Several recent methods study distributions specifically in the low-rank adaptation space. ICM-LoRA learns a conditional generative model from a collection of LoRAs that were optimized beforehand~\citep{shao2025icmlora}. DiffLoRA similarly learns to generate LoRA weights from previously constructed personalization adapters using a diffusion model~\citep{wu2024difflora}. These methods provide important evidence that LoRA weights can be modeled with a conditional generative distribution, but the distribution is learned by reproducing previously optimized adapters. Bayesian LoRA methods take a different view. Laplace-LoRA constructs a distribution around a fitted LoRA after adaptation, while BLoB learns a variational distribution over LoRA parameters during fine-tuning~\citep{yang2023laplacelora,wang2024blob}. Amortized Bayesian meta-learning similarly represents uncertainty over task-specific low-rank adaptations~\citep{zhang2025abmll}. LoRA ensembles obtain weight diversity by independently optimizing several adapters for the same task~\citep{wang2023loraensembles}.

Our distribution is different in both how it is produced and how it is used. It is conditioned on each query and trained directly through the frozen language model's response loss rather than by reconstructing a set of adapters or approximating a task-level posterior. Samples from the distribution are treated as alternative adaptations for the current query rather than only as uncertainty estimates. This also means that the predicted distribution may change from one query to the next, even when the queries come from the same benchmark or task.

\paragraph{Test-Time Scaling.}
Most test-time scaling methods keep the model parameters fixed and spend additional compute by generating and selecting among more output sequences~\citep{snell2024scaling,shao2024deepseekmath,muennighoff2025s1}. Other work shows that useful variation can also exist in weight space. AdapterSoup and Model Soups average independently trained parameters~\citep{chronopoulou2023adaptersoup,wortsman2022model}, while LoRAHub, LoRARetriever, and LoRA-Flow retrieve or combine adapters for a new input or task~\citep{huang2023lorahub,zhao2024loraretriever,wang2024loraflow}. Model Swarms searches among and modifies model experts during inference, and evolutionary weight perturbations provide another way to search for diverse solutions in parameter space~\citep{feng2024modelswarms,hayes2026beyond}. These results motivate weight space itself as a useful source of test-time diversity.

Our approach learns this search distribution during training. At test time, the hypernetwork directly produces a distribution from which multiple adapted models can be sampled. We study this in two settings. The regression hypernetwork predicts a distribution directly over LoRA weights, while the mixing hypernetwork predicts a distribution over combinations of reference LoRAs. The latter is related to adapter retrieval and composition, but instead of returning one composition, it defines several possible adaptations that can be explored with additional test-time compute.

\section{Hypernetwork Details}
\label{app:hypernet_details}
\subsection{Hypernetwork Architecture}
\label{app:hypernet_architecture}
\begin{figure}[t]
    \centering
    \includegraphics[width=0.45\linewidth]
    {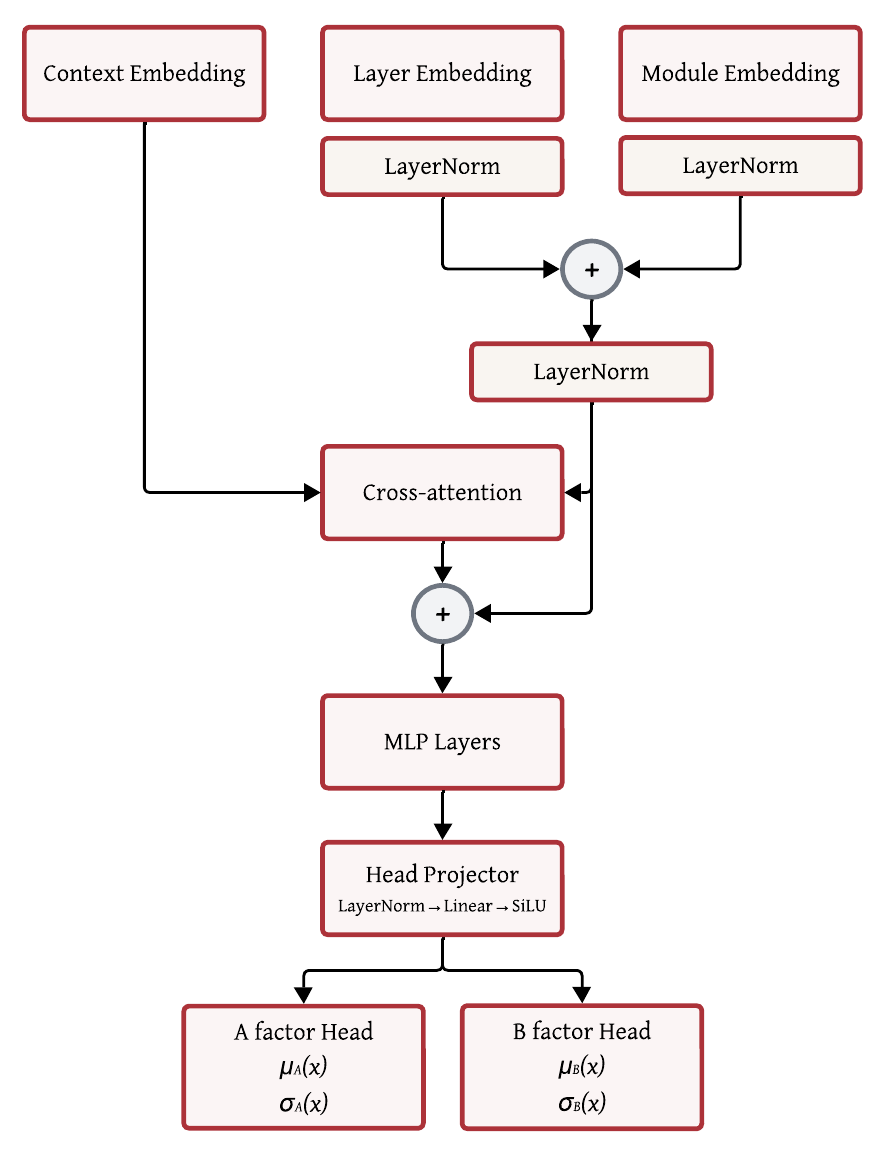}
    \caption{\textbf{Hypernetwork architecture.}
    Each layer--module pair attends over the token-level query representation and produces module-specific LoRA factor parameters.}
    \label{fig:hypernet_architecture}
\end{figure}

Figure~\ref{fig:hypernet_architecture} shows the architecture used for the hypernetwork. The hypernetwork is conditioned on three inputs: the token-level representation of the input query, a learned layer embedding, and a learned module embedding. The query is first passed through the frozen language model to obtain a sequence of hidden states. The hidden states are projected to the internal dimension of the hypernetwork before being used as the context representation.

We use this pair representation as the \emph{query} to a cross-attention layer, while the sequence of query token representations acts as the keys and values. Thus, each $(\text{layer},\text{module})$ pair can attend to a different part of the input when constructing its LoRA parameters. The cross-attention output is combined with the pair representation through a residual connection and then processed by eight residual MLP blocks. A final head projector consisting of LayerNorm, a linear layer, and a SiLU activation produces the representation used by the output heads.

The output layer contains separate heads for the two LoRA factors $\text{A}$ and $\text{B}$. For a distributional version of the hypernetwork with diagonal covariance, the same features are used to predict both the mean and element-wise standard deviation of each factor, $\mu_\text{A}(x), \sigma_\text{A}(x), \mu_\text{B}(x)$, and $\sigma_\text{B}(x)$. A sampled LoRA is obtained through the reparameterization
\begin{equation}
\text{A} = \mu_\text{A}(x) + \sigma_\text{A}(x)\odot\epsilon_\text{A}, \qquad \epsilon_\text{A} \sim \mathcal{N}(0,I),
\end{equation}
\begin{equation}
\text{B} = \mu_\text{B}(x) + \sigma_\text{B}(x)\odot\epsilon_\text{B}, \qquad \epsilon_\text{B} \sim \mathcal{N}(0,I), 
\end{equation}
where the noise for $\text{A}$ and $\text{B}$ is sampled independently. The deterministic hypernetwork uses the same architecture but predicts only the corresponding $\text{A}$ and $\text{B}$ point estimates.

\subsection{Training Details}
\label{app:training_details}
\begin{figure}[t]
    \centering
    \includegraphics[width=0.85\linewidth]{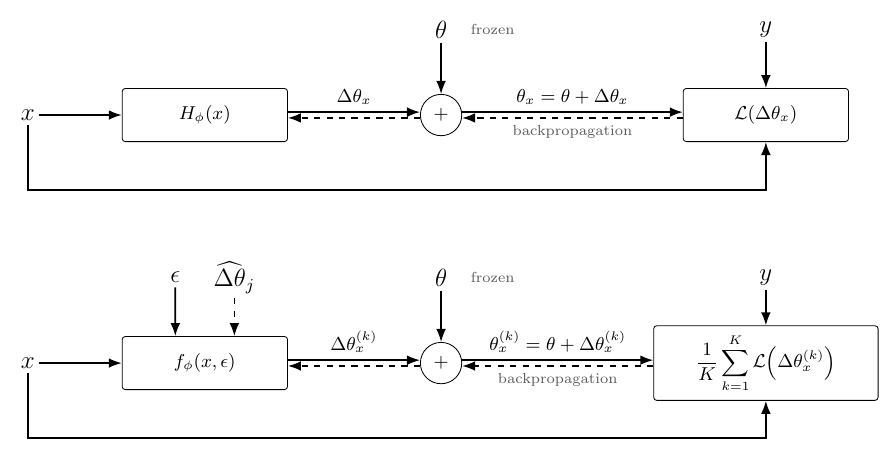}
    \caption{
    \textbf{Training the hypernetwork.}
    The hypernetwork generates an update $\Delta W_x$, which is added to the frozen language model weights $W_0$. The language modeling loss is backpropagated through the adapted model and the generated update to train only the hypernetwork parameters.}
    \label{fig:hypernet_training}
\end{figure}
For the distributional hypernetwork, we sample $K$ candidate LoRA updates for each query and evaluate the language modeling loss for each corresponding adapted model. Under the best-of-$K$ objective, the candidate with the lowest loss is selected, and the backward pass is performed through that sampled update. Figure~\ref{fig:hypernet_training} illustrates this gradient flow using the colored arrows. The gradient passes from the language modeling loss through the adapted language model, through the generated LoRA update, and finally into the hypernetwork that produced it. Although gradients are propagated through the language model to obtain the learning signal, its pretrained weights remain frozen throughout training. Only the hypernetwork parameters are updated. Because the sampled LoRA is produced through the reparameterized distribution, the same gradient also trains the parameters controlling both the center and spread of the predicted update distribution.

\subsection{Hypernetwork Parameter Count}
\label{app:hypernet_size}
The size of the hypernetwork depends on the base language model. Although we use rank-$8$ LoRA updates throughout the main experiments, different base models have different hidden dimensions and numbers and shapes of the target matrices. Consequently, the number of LoRA parameters that the hypernetwork must generate varies across models. Since the output heads of the hypernetwork scale with the dimensionality of these generated LoRA factors, the total hypernetwork parameter count varies accordingly.

Table~\ref{tab:hypernet_size} reports the parameter counts for each base model. The deterministic hypernetwork predicts one set of LoRA factors, whereas the distributional hypernetwork additionally predicts the parameters controlling the spread of the distribution. For the diagonal Gaussian parameterization used in our main experiments, this introduces additional $\sigma$ heads for the LoRA factors and therefore increases the size of the hypernetwork.

We also examine how the parameter count changes with the LoRA rank. Table~\ref{tab:rank_parameter_scaling} shows this scaling for Qwen3-1.7B and Qwen3-4B, comparing FullFT, standard LoRA, and the deterministic hypernetwork. As the rank increases, both the generated LoRA update and the hypernetwork output heads grow accordingly.

\begin{table*}[t]
\centering
\caption{
Parameter counts across base models and adaptation methods.
FullFT reports the number of trainable LM parameters, while LoRA reports the
trainable parameters of a single rank-$8$ adapter.
For mixing hypernetworks, the reported count includes both the hypernetwork and
the fixed bank of 50 rank-$8$ reference LoRAs.
}
\label{tab:hypernet_size}
\small
\setlength{\tabcolsep}{5pt}
\renewcommand{\arraystretch}{1.08}
\resizebox{\textwidth}{!}{%
\begin{tabular}{lcc cc cc}
\toprule
&
&
&
\multicolumn{2}{c}{\textbf{Regression Hypernet}}
&
\multicolumn{2}{c}{\textbf{Mixing Hypernet + Bank}} \\
\cmidrule(lr){4-5}
\cmidrule(lr){6-7}
\textbf{Base model}
& \textbf{FullFT}
& \textbf{LoRA $r=8$}
& \textbf{Det.}
& \textbf{Dist.}
& \textbf{Det.}
& \textbf{Dist.} \\
\midrule
Qwen3-1.7B
& 1,700M
& 5.5M
& 194M
& 295M
& 368.0M
& 368.1M \\

Qwen3-4B
& 4,000M
& 10.6M
& 245M
& 396M
& 624.1M
& 624.2M \\

Gemma-3-1B
& 1,000M
& 5.0M
& 192M
& 291M
& 343.3M
& 343.4M \\

Llama-3.2-1B
& 1,000M
& 3.9M
& 219M
& 345M
& 289.3M
& 289.4M \\

Llama-3.2-3B
& 3,000M
& 7.6M
& 233M
& 372M
& 472.4M
& 472.5M \\
\bottomrule
\end{tabular}%
}
\end{table*}

\begin{table}[t]
\centering
\caption{
Parameter scaling with LoRA rank for Qwen3-1.7B and Qwen3-4B.
The LoRA row gives the number of trainable parameters for standard LoRA,
which is also the size of the update generated by the hypernetwork.
}
\label{tab:rank_parameter_scaling}
\small
\setlength{\tabcolsep}{7pt}
\renewcommand{\arraystretch}{1.08}

\begin{tabular}{lcccc}
\toprule
\multicolumn{5}{c}{\textbf{Qwen3-1.7B}} \\
\midrule
\textbf{Approach}
& \textbf{Rank 2}
& \textbf{Rank 8}
& \textbf{Rank 32}
& \textbf{Rank 128} \\
\midrule
FullFT trainable
& $\sim$1,700M
& $\sim$1,700M
& $\sim$1,700M
& $\sim$1,700M \\

LoRA trainable / generated update
& 1.4M
& 5.5M
& 22.0M
& 88.1M \\

Deterministic hypernet trainable
& 118M
& 194M
& 497M
& 1,707M \\

\midrule
\multicolumn{5}{c}{\textbf{Qwen3-4B}} \\
\midrule
\textbf{Approach}
& \textbf{Rank 2}
& \textbf{Rank 8}
& \textbf{Rank 32}
& \textbf{Rank 128} \\
\midrule
FullFT trainable
& $\sim$4,000M
& $\sim$4,000M
& $\sim$4,000M
& $\sim$4,000M \\

LoRA trainable / generated update
& 2.7M
& 10.6M
& 42.5M
& 169.9M \\

Deterministic hypernet trainable
& 132M
& 245M
& 699M
& 2,514M \\

\bottomrule
\end{tabular}
\end{table}

\section{Implementation Details}
\label{app:Implementation}
\paragraph{Sampling.}
During training, we draw Gaussian noise in antithetic pairs, $\epsilon$ and $-\epsilon$, to reduce the variance introduced by sampling.

\paragraph{Implementation of the mixture parameterization.}
Equation~\ref{eq:mixture} sums $\text{R}$ full-size updates, which would be wasteful to
form explicitly. We instead evaluate it as a single low-rank product. For each target
matrix $s\in\mathcal{M}$, write the reference adapters in factored form
$\widehat{\Delta\theta}_j^{(s)}=\frac{\alpha}{r}\text{B}_j^{(s)}\text{A}_j^{(s)}$ with
$\text{A}_j^{(s)}\in\mathbb{R}^{r\times d_{\mathrm{in}}}$ and
$\text{B}_j^{(s)}\in\mathbb{R}^{d_{\mathrm{out}}\times r}$, and stack them, scaling only
the $\text{A}$ side by the sampled mixing weights:
\begin{equation}
\text{A}_{\mathrm{cat}}^{(s)}
=
\begin{bmatrix}
\pi_{x,1}^{(k)}\text{A}_1^{(s)}\\[-1pt] \vdots \\[-1pt] \pi_{x,\text{R}}^{(k)}\text{A}_\text{R}^{(s)}
\end{bmatrix},
\qquad
\text{B}_{\mathrm{cat}}^{(s)}
=
\bigl[\,\text{B}_1^{(s)} \;\cdots\; \text{B}_\text{R}^{(s)}\,\bigr],
\qquad
\text{B}_{\mathrm{cat}}^{(s)}\text{A}_{\mathrm{cat}}^{(s)}
=
\sum_{j=1}^{\text{R}} \pi_{x,j}^{(k)}\,\text{B}_j^{(s)}\text{A}_j^{(s)}
\label{eq:cat}
\end{equation}
The right-hand identity is exact, not an approximation, and holds for any $\pi$, we verify it numerically to floating-point tolerance. The sampled adaptation is therefore applied as one LoRA of rank $\text{R}r$ rather than as $\text{R}$ separate updates, and by linearity it
is also equivalent to applying each reference adapter to the input and summing the outputs
with weights $\pi_x^{(k)}$, as mixture-of-LoRA architectures do. Two consequences are worth
stating. First, the effective rank of a mixture sample is up to $\text{R}r$, whereas
averaging the factors instead of the updates would keep rank $r$; the latter is the
convention used for \emph{merging} adapters into a single deployable module
(\texttt{peft}'s \texttt{linear} mode, LoRAhub), but it computes
$(\sum_j\pi_j\text{B}_j)(\sum_j\pi_j\text{A}_j)$, which differs from
Equation~\ref{eq:mixture} by cross terms $\text{B}_i\text{A}_j$, $i\neq j$, that carry
$1-\sum_j\pi_j^2$ of the total weight and were never trained. Since we require samples to be
convex combinations of the reference updates, we use the exact form throughout. Second, this
costs no additional parameters: $\text{A}_{\mathrm{cat}}$ and $\text{B}_{\mathrm{cat}}$ are
views of the frozen bank, and only the adapter-path compute grows with $\text{R}$.

\section{Additional Experiments}

\subsection{Effect of Hypernetwork-generated LoRA Placement}
\label{app:lora_placement}
\begin{figure}[t]
    \centering
    \includegraphics[width=0.55\linewidth]
    {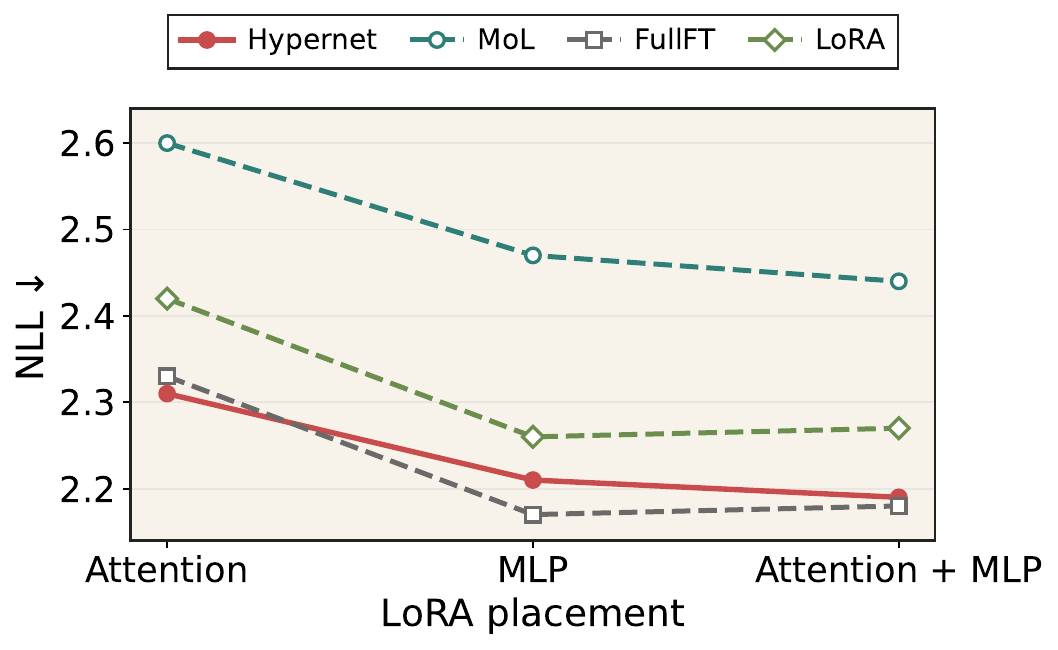}
    \caption{Effect of LoRA placement. Hypernetwork-generated updates follow the same trend as directly optimized LoRAs: adapting the MLP is more effective than adapting attention alone, while adapting both provides only a small additional improvement.}
    \label{fig:lora_placement}
\end{figure}

We additionally study where the hypernetwork-generated LoRA update should be applied. Figure~\ref{fig:lora_placement} compares adapting the attention projections, the MLP projections, or both. The generated updates follow the same qualitative behavior as directly optimized LoRAs. Updating the MLP gives a larger improvement than updating attention alone, while adapting both attention and MLP provides only a small additional gain. This suggests that the behavior of the generated updates is consistent with ordinary gradient-based LoRA adaptation rather than relying on a particular choice of target module.

\subsection{Effect of Hypernetwork Capacity.}
\label{app:hypernet_capacity}
We vary the size of the distributional hypernetwork from approximately
25M to 1B parameters while keeping the base model and training objective fixed. Larger hypernetworks generally improve performance, but the benefit is much more pronounced when using a single predicted update. For Qwen3-4B, for example, increasing the hypernetwork size from 25M to 1B improves accuracy by 4.5 points at $K=1$, but by only 0.9 points at $K=64$. A similar pattern appears for Qwen3-1.7B.

These results suggest that hypernetwork capacity is most important for
producing a strong individual update, while sampling multiple weight updates reduces the dependence on having a very large hypernetwork. Since our main setting uses the distribution specifically for test-time weight sampling, we use moderately sized hypernetworks in the main experiments rather than scaling the hypernetwork aggressively.

\begin{table}[t]
\centering
\caption{Effect of hypernetwork capacity on distributional weight sampling. We report accuracy for different hypernetwork sizes and sampling budgets. Increasing hypernetwork capacity generally improves single-sample performance, while the gap between small and large hypernetworks becomes much smaller as the number of sampled weight updates increases.}
\label{tab:hypernet_capacity}
\small
\begin{tabular}{lcccc}
\toprule
\textbf{Base model} & \textbf{Hypernet size} & \textbf{$K=1$} & \textbf{$K=8$} & \textbf{$K=64$} \\
\midrule
\multirow{4}{*}{Qwen3-1.7B}
& 25M  & 44.7 & 51.9 & 53.3 \\
& 100M & 45.3 & 53.0 & 53.7 \\
& 500M & 44.5 & 53.2 & 53.9 \\
& 1B   & 46.7 & 53.5 & 54.2 \\
\midrule
\multirow{4}{*}{Qwen3-4B}
& 25M  & 51.3 & 60.1 & 61.2 \\
& 100M & 50.1 & 60.5 & 61.3 \\
& 500M & 54.1 & 61.4 & 61.8 \\
& 1B   & 55.8 & 61.4 & 62.1 \\
\bottomrule
\end{tabular}
\end{table}

\section{When Does Generating an Update Become Useful?}
\label{sec:when_generation}

\begin{figure}
    \centering
    \vspace{-1em}
    \includegraphics[width=0.55\linewidth]{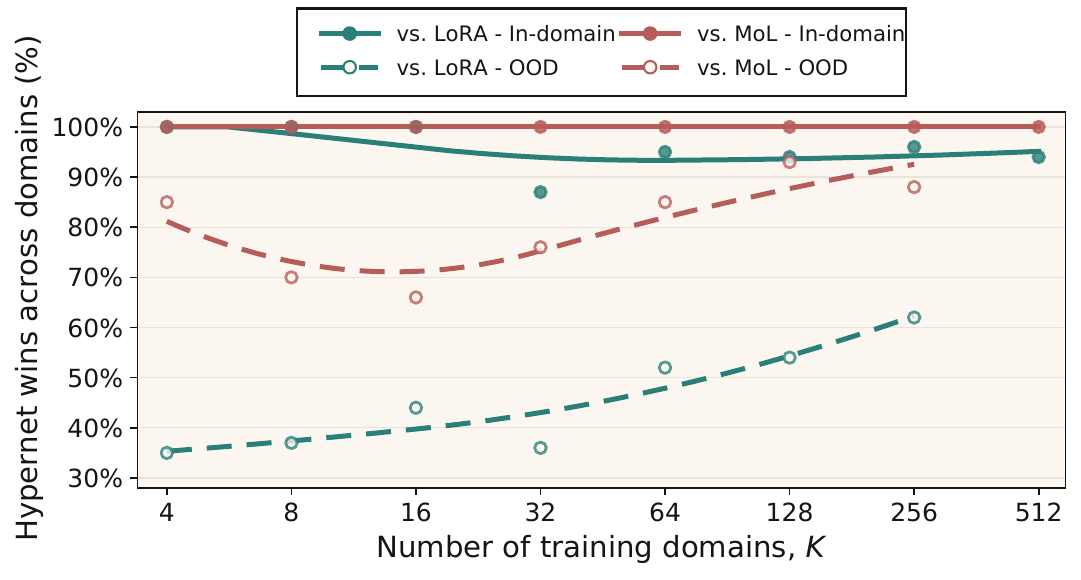}
    \caption{\textbf{The benefit of generating updates grows with task diversity.} We report the percentage of domains where the hypernetwork outperforms each baseline. Higher values indicate a larger advantage for the hypernetwork.}
    \label{fig:diversity_generation}
\end{figure}
Generating a different update for every query may not always be necessary. If different inputs require similar adaptations, a single shared LoRA may be sufficient. We therefore vary the diversity of the training distribution by training on $K$ domains, with $K$ ranging from 4 to 512, and compare the deterministic hypernetwork with a single LoRA and MoL.

Figure~\ref{fig:diversity_generation} shows that the benefit of generation increases with task diversity. On in-domain data, the hypernetwork consistently outperforms the shared LoRA. On unseen domains, the shared LoRA remains competitive at low diversity, but becomes less effective as $K$ increases and a single update becomes less representative of what different queries require. MoL provides a stronger query-conditioned baseline, but remains limited to selecting from a fixed set of updates, while the hypernetwork can generate a new update for each query.

\begin{paperbox}{Finding 4: \textit{Task Diversity Determines When Updates Should Be Generated}}
A shared update works well when tasks are similar, but becomes less representative as task diversity grows. At high diversity, generating input-conditioned updates scales better than relying on a fixed update or a fixed bank of updates.
\end{paperbox}

\section{Learning a Distribution over Weight Updates}
\label{sec:learning_dist_app}

\subsection{Choosing the Covariance}
\label{sec:covariance}

A distributional hypernetwork also requires a choice of how to represent the covariance over its generated updates. We compare diagonal, block diagonal, and full covariance parameterizations. Since directly predicting the latter two would be expensive, we represent them using low-rank factorizations.

Figure~\ref{fig:distribution_design}(left) shows that a more expressive covariance provides little benefit. The three parameterizations perform similarly at different sampling budgets, while the diagonal covariance performs slightly better as the number of samples increases. At $K=8$, for example, the diagonal covariance reaches approximately 0.75 mean accuracy, compared with 0.74 for block diagonal and 0.68 for full covariance. We therefore use a diagonal covariance throughout the remaining experiments, since it is both simpler and cheaper to parameterize. The results shown are mean accuracy across 8 benchmarks.

\subsection{Learning a Searchable Distribution}
\label{sec:searchable_distribution}

The distribution is useful at test-time only if different samples continue to provide meaningfully different candidate updates. This creates a second requirement on the training objective. If the loss requires a large fraction of the sampled updates to perform well, the hypernetwork can reduce the loss by shrinking the predicted variance and concentrating its samples around a single update. We refer to this behavior as \emph{distribution collapse}, where the predicted variance shrinks and repeated samples approach the same update.


We therefore compare objectives that place different amounts of pressure on the sampled distribution. The mean objective optimizes all sampled updates, while the median and top fraction objectives require progressively smaller portions of the distribution to perform well. Best-of-$K$ places the weakest constraint on the full distribution, since only the best sampled update receives the optimization signal.

The choice of $K$ also controls the balance between guiding the center of the distribution and preserving its spread. With a small finite $K$, the best sample remains closely coupled to the mean, providing a useful signal for where the mean should move. As $K$ increases, the objective becomes more tolerant of a wider distribution, since a good sample can be found farther from the mean. Top fraction objectives provide an intermediate tradeoff, applying more pressure on the mean and favoring a smaller variance than best-of-$K$, while still avoiding the stronger contraction encouraged by the mean objective.

We measure the resulting coverage using oracle performance as the number of weight samples increases. As shown in Figure~\ref{fig:distribution_design}(right), the difference becomes clear as the sampling budget grows. Thus, objectives that require more of the distribution to perform well leave substantially less room for search, while best-of-$K$ retains the largest searchable region.

Mode collapse is a distinct failure mode in generative modeling, where a learned distribution fails to represent multiple valid modes~\citep{soletskyi2025theoretical}. Our concern here is different: the predicted variance may shrink toward zero. Mode collapse concerns whether a model covers several distinct valid target modes. Distribution collapse instead refers to the predicted variance shrinking toward zero, so that repeated samples produce the same update. Our goal here is not to guarantee coverage of every possible mode, but to retain variation in weight space for sampling to remain useful at test-time.

\begin{paperbox}{Finding 5: \textit{Best-of-$K$ Loss Preserves a Searchable Weight Distribution}}
Objectives that require many sampled updates to perform well encourage the learned distribution to contract. Best-of-$K$ places less pressure on the full distribution and retains substantially more coverage as the weight sampling budget increases.
\end{paperbox}

\section{Full Experimental Results}

\subsection{Benchmark-wise Comparison Between Deterministic and Distributional Hypernetworks}
\label{app:dist_vs_det_full}

Table~\ref{tab:dist_train_benchmark_full} reports the full benchmark-wise results corresponding to the comparison in Section~\ref{sec:distribution_improves_update}. Across the five models, distributional training generally improves macro performance over a directly trained deterministic hypernet.

\begin{table*}[t]
\centering
\scriptsize
\setlength{\tabcolsep}{2.2pt}
\renewcommand{\arraystretch}{1.08}
\caption{Benchmark-level comparison of deterministic and distributional training for regression and mixing hypernetworks across five base models on the seven benchmarks used in our main evaluation. Accuracy is reported in percentage points. Higher is better.}
\label{tab:dist_train_benchmark_full}
\resizebox{\textwidth}{!}{%
\begin{tabular}{lcccccccccccccccccccc}
\toprule
& \multicolumn{4}{c}{Qwen3 1.7B} & \multicolumn{4}{c}{Qwen3 4B} & \multicolumn{4}{c}{Gemma 3 1B} & \multicolumn{4}{c}{Llama 3.2 1B} & \multicolumn{4}{c}{Llama 3.2 3B} \\
\cmidrule(lr){2-5}\cmidrule(lr){6-9}\cmidrule(lr){10-13}\cmidrule(lr){14-17}\cmidrule(lr){18-21}
& \multicolumn{2}{c}{Regression} & \multicolumn{2}{c}{Mixing} & \multicolumn{2}{c}{Regression} & \multicolumn{2}{c}{Mixing} & \multicolumn{2}{c}{Regression} & \multicolumn{2}{c}{Mixing} & \multicolumn{2}{c}{Regression} & \multicolumn{2}{c}{Mixing} & \multicolumn{2}{c}{Regression} & \multicolumn{2}{c}{Mixing} \\
\cmidrule(lr){2-3}\cmidrule(lr){4-5}\cmidrule(lr){6-7}\cmidrule(lr){8-9}\cmidrule(lr){10-11}\cmidrule(lr){12-13}\cmidrule(lr){14-15}\cmidrule(lr){16-17}\cmidrule(lr){18-19}\cmidrule(lr){20-21}
Benchmark & Det. & Dist. & Det. & Dist. & Det. & Dist. & Det. & Dist. & Det. & Dist. & Det. & Dist. & Det. & Dist. & Det. & Dist. & Det. & Dist. & Det. & Dist. \\
\midrule
GSM8K & 56.8 & \textbf{64.0} & \textbf{78.0} & 69.2 & \textbf{64.4} & 62.4 & \textbf{88.0} & 82.4 & \textbf{29.2} & 28.8 & 30.4 & \textbf{41.6} & \textbf{44.0} & 40.8 & 49.2 & \textbf{52.0} & 68.8 & \textbf{72.4} & 82.8 & \textbf{84.4} \\
MATH500 & 34.4 & \textbf{42.0} & 46.0 & \textbf{47.2} & 52.8 & \textbf{56.0} & \textbf{56.4} & 45.2 & 20.0 & \textbf{21.2} & 18.4 & \textbf{26.4} & 22.0 & \textbf{23.6} & \textbf{24.8} & 23.2 & \textbf{39.2} & 38.0 & \textbf{40.4} & 37.6 \\
HumanEval & 45.1 & \textbf{47.6} & 48.2 & \textbf{59.2} & 59.2 & \textbf{61.0} & \textbf{66.5} & 65.2 & 7.9 & \textbf{12.8} & 15.8 & \textbf{24.4} & 30.5 & \textbf{31.7} & 36.0 & \textbf{38.4} & 45.7 & \textbf{48.8} & 48.8 & \textbf{51.2} \\
ARC-C & 59.2 & \textbf{68.0} & 71.6 & \textbf{76.0} & \textbf{86.0} & 85.2 & 83.6 & \textbf{86.4} & 23.6 & \textbf{30.4} & 22.0 & \textbf{43.6} & 19.6 & \textbf{27.2} & 50.0 & \textbf{54.8} & 65.2 & \textbf{72.8} & \textbf{76.0} & 74.4 \\
MMLU-STEM & 52.0 & \textbf{64.8} & 58.8 & \textbf{66.8} & \textbf{76.4} & 74.4 & \textbf{82.0} & 81.2 & 23.6 & \textbf{25.2} & 20.0 & \textbf{40.0} & 17.2 & \textbf{28.8} & 34.8 & \textbf{39.6} & 47.2 & \textbf{48.8} & \textbf{61.2} & 58.8 \\
MedQA & 23.6 & \textbf{32.4} & 27.6 & \textbf{44.0} & 55.2 & \textbf{60.4} & 57.2 & \textbf{60.4} & 9.2 & \textbf{15.2} & 15.2 & \textbf{21.6} & 18.0 & \textbf{23.2} & 40.8 & \textbf{41.6} & 49.2 & \textbf{59.6} & \textbf{62.0} & 60.0 \\
GPQA-Diamond & 17.7 & \textbf{19.7} & 20.7 & \textbf{27.8} & \textbf{35.9} & 31.3 & 35.9 & \textbf{36.4} & 10.6 & \textbf{14.1} & 14.6 & \textbf{21.7} & 11.6 & \textbf{14.1} & 18.2 & \textbf{22.7} & \textbf{16.7} & 13.1 & \textbf{25.2} & \textbf{25.2} \\
\midrule
\textbf{Macro} & 41.3 & \textbf{48.4} & 50.1 & \textbf{55.7} & 61.4 & \textbf{61.5} & \textbf{67.1} & 65.3 & 17.7 & \textbf{21.1} & 19.5 & \textbf{31.3} & 23.3 & \textbf{27.1} & 36.3 & \textbf{38.9} & 47.4 & \textbf{50.5} & \textbf{56.6} & 56.0 \\
\bottomrule
\end{tabular}%
}
\end{table*}

\subsection{Full Benchmark-Level Test-Time Scaling Results}
\label{app:full_test_time_scaling}

Table~\ref{tab:benchmark_scaling_summary} summarizes the benchmark-level results underlying Figure~\ref{fig:weight_sampling}. To keep the appendix
compact, we report representative low, intermediate, and high sampling budgets, $K\in\{1,8,32\}$, while the main figure shows the full scaling curves. For deterministic methods, additional samples are generated from fixed adapted weights. For distributional hypernet, each generation uses a newly sampled weight update. Results are reported on GSM8K, MATH500, HumanEval, ARC-Challenge, MMLU-STEM, MedQA, and GPQA-Diamond, together with their macro average.

As in the main text, regression and mixing hypernetwork are not compared directly. The reference LoRAs used by the mixing hypernetwork are trained for three epochs before one epoch of hypernetwork training, whereas the regression hypernetwork is trained for one epoch on the same data.

\begin{table*}[t]
\centering
\scriptsize
\setlength{\tabcolsep}{2.8pt}
\renewcommand{\arraystretch}{0.94}
\caption{\textbf{Benchmark-level test-time scaling summary.} Each cell reports accuracy at $K=1/8/32$. $\mathrm{T}$ denotes token sampling from fixed adapted weights and $\mathrm{W}$ denotes sampling a new weight update for each generation. Regression and mixing hypernetwork should be interpreted separately because their training exposure is not matched.}
\label{tab:benchmark_scaling_summary}
\resizebox{\textwidth}{!}{%
\begin{tabular}{@{}llc*{8}{c}@{}}
\toprule
\textbf{Model} & \textbf{Method} & \textbf{Samp.} & \textbf{GSM8K} & \textbf{MATH} & \textbf{HumanEval} & \textbf{ARC-C} & \textbf{MMLU} & \textbf{MedQA} & \textbf{GPQA} & \textbf{Macro} \\
\midrule
\multirow{7}{*}{\textbf{Qwen3 1.7B}} & Full FT & T & .508/.700/.768 & .316/.444/.540 & .529/.643/.714 & .608/.740/.788 & .496/.664/.700 & .272/.424/.428 & .217/.273/.354 & .421/.555/.613 \\
 & LoRA & T & .568/.620/.632 & .316/.428/.500 & .667/.607/.635 & .520/.760/.800 & .492/.664/.708 & .172/.336/.372 & .177/.283/.268 & .416/.528/.559 \\
 & MoL & T & .468/.560/.608 & .292/.428/.480 & .707/.701/.739 & .596/.704/.740 & .512/.688/.684 & .112/.320/.352 & .202/.293/.263 & .413/.528/.552 \\
\cmidrule(lr){2-11}
 & Reg. Det. & T & .524/.588/.612 & .312/.420/.504 & .633/.702/.738 & .528/.736/.748 & .496/.676/.688 & .196/.356/.368 & .207/.303/.308 & .414/.540/.567 \\
 & \textbf{Reg. Dist.} & W & .620/.684/.740 & .396/.528/.556 & .688/.705/.706 & .708/.760/.784 & .608/.692/.720 & .308/.444/.428 & .197/.278/.258 & .504/.584/.599 \\
\cmidrule(lr){2-11}
 & Mix. Det. & T & .776/.844/.864 & .456/.528/.552 & .739/.781/.791 & .648/.748/.752 & .608/.744/.760 & .280/.388/.416 & .212/.298/.293 & .531/.619/.633 \\
 & \textbf{Mix. Dist.} & W & .624/.720/.736 & .416/.540/.536 & .737/.765/.771 & .752/.788/.776 & .668/.744/.744 & .404/.432/.424 & .253/.298/.283 & .551/.612/.610 \\
\addlinespace[0.2em]
\midrule
\addlinespace[0.1em]
\multirow{7}{*}{\textbf{Qwen3 4B}} & Full FT & T & .680/.812/.856 & .496/.620/.656 & .706/.731/.754 & .832/.884/.912 & .736/.796/.808 & .504/.584/.600 & .318/.338/.338 & .610/.681/.703 \\
 & LoRA & T & .668/.704/.692 & .400/.564/.620 & .772/.786/.829 & .800/.904/.904 & .652/.804/.812 & .340/.552/.624 & .293/.333/.328 & .561/.664/.687 \\
 & MoL & T & .660/.720/.720 & .428/.564/.624 & .846/.828/.843 & .744/.876/.904 & .652/.780/.816 & .384/.560/.612 & .217/.348/.374 & .562/.668/.699 \\
\cmidrule(lr){2-11}
 & Reg. Det. & T & .660/.700/.708 & .436/.548/.600 & .755/.815/.841 & .764/.884/.884 & .692/.816/.792 & .508/.604/.600 & .318/.384/.359 & .590/.679/.683 \\
 & \textbf{Reg. Dist.} & W & .628/.660/.660 & .496/.604/.624 & .792/.839/.820 & .804/.876/.868 & .724/.788/.772 & .536/.580/.588 & .293/.359/.348 & .610/.672/.669 \\
\cmidrule(lr){2-11}
 & Mix. Det. & T & .880/.928/.928 & .568/.628/.664 & .808/.846/.841 & .860/.860/.864 & .788/.816/.828 & .572/.580/.608 & .343/.364/.338 & .688/.717/.724 \\
 & \textbf{Mix. Dist.} & W & .832/.916/.928 & .424/.512/.552 & .860/.833/.845 & .888/.892/.888 & .752/.796/.804 & .568/.584/.592 & .354/.374/.369 & .668/.701/.711 \\
\addlinespace[0.2em]
\midrule
\addlinespace[0.1em]
\multirow{7}{*}{\textbf{Gemma 3 1B}} & Full FT & T & .240/.420/.516 & .120/.228/.276 & .568/.543/.600 & .220/.400/.392 & .192/.336/.364 & .060/.248/.276 & .106/.232/.258 & .215/.344/.383 \\
 & LoRA & T & .192/.388/.512 & .100/.184/.272 & .524/.491/.507 & .128/.336/.464 & .152/.300/.356 & .076/.216/.272 & .071/.182/.247 & .178/.300/.376 \\
 & MoL & T & .188/.300/.396 & .116/.212/.304 & .548/.463/.486 & .240/.428/.488 & .180/.336/.356 & .072/.240/.268 & .146/.268/.222 & .213/.321/.360 \\
\cmidrule(lr){2-11}
 & Reg. Det. & T & .212/.368/.460 & .132/.220/.292 & .633/.523/.493 & .152/.352/.472 & .144/.300/.328 & .096/.264/.292 & .106/.207/.242 & .211/.319/.368 \\
 & \textbf{Reg. Dist.} & W & .284/.428/.440 & .164/.228/.284 & .433/.500/.551 & .304/.464/.472 & .224/.356/.372 & .120/.260/.264 & .177/.222/.258 & .244/.351/.377 \\
\cmidrule(lr){2-11}
 & Mix. Det. & T & .300/.436/.500 & .116/.212/.260 & .386/.377/.443 & .148/.320/.396 & .188/.304/.356 & .140/.308/.292 & .217/.268/.253 & .214/.318/.357 \\
 & \textbf{Mix. Dist.} & W & .364/.480/.532 & .244/.308/.360 & .413/.500/.529 & .336/.492/.500 & .292/.388/.380 & .144/.300/.300 & .182/.242/.278 & .282/.387/.411 \\
\bottomrule
\end{tabular}%
}
\end{table*}

\end{document}